\documentclass{article}

\usepackage{PRIMEarxiv}

\usepackage[utf8]{inputenc}
\usepackage[T1]{fontenc}
\usepackage{hyperref}
\usepackage{url}
\usepackage{booktabs}
\usepackage{graphicx}
\usepackage{multirow}
\usepackage{enumitem}
\usepackage[table]{xcolor}
\usepackage{amsfonts}
\usepackage{amsmath}
\usepackage{nicefrac}
\usepackage{microtype}
\usepackage{fancyhdr}
\usepackage{graphicx}
\usepackage[most]{tcolorbox}
\usepackage{xcolor}
\usepackage{xspace}
\usepackage{wrapfig}
\graphicspath{{media/}}
\usepackage{etoolbox}
\usepackage{fontawesome5}
\usepackage{threeparttable}
\usepackage{eso-pic}

\usepackage{makecell}
\usepackage[normalem]{ulem}

\usepackage{tabularx}
\usepackage{siunitx}
\usepackage{array}
\usepackage{multicol}
\usepackage{enumitem}

\usepackage{CJKutf8}
\usepackage{ragged2e}
\usepackage{url}
\usepackage[labelfont=bf]{caption}
\usepackage{pdflscape}
\definecolor{asiA}{HTML}{D946EF}
\definecolor{asiS}{HTML}{6366F1}
\definecolor{asiI}{HTML}{06B6D4}
\definecolor{guidanceBoxBg}{HTML}{FBF7F7}
\definecolor{guidanceBoxAccent}{HTML}{981B14}

\usepackage{booktabs}
\usepackage{multirow}
\usepackage[table]{xcolor}

\definecolor{MainPurple}{HTML}{665BD6}
\definecolor{BOne}{HTML}{6B60D8}
\definecolor{BTwo}{HTML}{DC9849}
\definecolor{BThree}{HTML}{617FD8}
\definecolor{BFour}{HTML}{DDB937}

\definecolor{RuleGray}{HTML}{A9ACB5}
\arrayrulecolor{RuleGray}

\definecolor{abstractBoxBg}{HTML}{EEF6FF}
\definecolor{abstractBoxAccent}{HTML}{2563A8}
\definecolor{projectLinkBlue}{HTML}{174A8B}

\newcolumntype{Y}{>{\centering\arraybackslash}X}

\usepackage{booktabs}
\usepackage{tabularx}
\usepackage{array}
\usepackage{threeparttable}
\usepackage{multirow}
\usepackage[table]{xcolor}

\newcolumntype{Y}{>{\centering\arraybackslash}X}

\usepackage{booktabs}
\usepackage{tabularx}
\usepackage{array}
\usepackage{threeparttable}
\usepackage{multirow}
\usepackage[table]{xcolor}

\newcolumntype{Y}{>{\centering\arraybackslash}X}

\newcommand{\taskSubmissionUrl}{https://asibench.apexin.ai/submit}

\newcommand{\projectWebsiteUrl}{https://asibench.apexin.ai/}
\newcommand{\projectGithubUrl}{https://github.com/apexin-ai/ASI-Bench.git}
\newcommand{\projectHuggingFaceUrl}{https://huggingface.co/datasets/Apexintelligence-AI/ASI-Bench-seed31415}

\newcommand{\projectLeaderboardUrl}{https://huggingface.co/spaces/Apexintelligence-AI/ASI-Bench-Leaderboard}

\newcommand{\projectlink}[3]{%
  \ifdefempty{#1}{%
    \textcolor{projectLinkBlue}{#2\,\textbf{#3}}%
  }{%
    \href{#1}{\textcolor{projectLinkBlue}{#2\,\textbf{#3}}}%
  }%
}

\newcommand{\asibench}{ASI-Bench\xspace}
\newcommand{\asiTitleA}{\raisebox{-0.08ex}{\scalebox{1.24}{\textbf{\textcolor{asiA}{A}}}}}
\newcommand{\asiTitleS}{\raisebox{-0.08ex}{\scalebox{1.24}{\textbf{\textcolor{asiS}{S}}}}}
\newcommand{\asiTitleI}{\raisebox{-0.08ex}{\scalebox{1.24}{\textbf{\textcolor{asiI}{I}}}}}

\usepackage[most]{tcolorbox}
\usepackage{xcolor}

\definecolor{taskblue}{RGB}{238,245,252}
\definecolor{taskblueframe}{RGB}{74,120,168}

\definecolor{b1blue}{RGB}{242,247,252}
\definecolor{b1frame}{RGB}{92,132,173}

\definecolor{b2blue}{RGB}{238,245,252}
\definecolor{b2frame}{RGB}{74,120,168}

\definecolor{b3blue}{RGB}{234,242,250}
\definecolor{b3frame}{RGB}{63,108,156}

\definecolor{b4blue}{RGB}{230,239,248}
\definecolor{b4frame}{RGB}{52,96,145}

\definecolor{guidanceBoxBg}{RGB}{238,245,252}
\definecolor{guidanceBoxAccent}{RGB}{74,120,168}

\renewcommand{\arraystretch}{1.0}

\newtcolorbox{taskbox}{
  enhanced,
  breakable,
  colback=taskblue,
  colframe=taskblueframe,
  boxrule=0.8pt,
  arc=2mm,
  left=2mm,
  right=2mm,
  top=1.5mm,
  bottom=1.5mm,
  fonttitle=\bfseries,
  title={Representative Scientific Task}
}

\newtcolorbox{bpromptbox}[3]{
  enhanced,
  breakable,
  colback=#2,
  colframe=#3,
  boxrule=0.8pt,
  arc=2mm,
  left=2mm,
  right=2mm,
  top=1.5mm,
  bottom=1.5mm,
  fonttitle=\bfseries,
  title={#1}
}

\newtcolorbox{guidancebox}{
  enhanced,
  breakable,
  colback=guidanceBoxBg,
  boxrule=0pt,
  sharp corners,
  left=1.1em,
  right=1.1em,
  top=0.7em,
  bottom=0.5em,
  before skip=0.8em,
  after skip=0.8em,
  borderline west={3pt}{0pt}{guidanceBoxAccent},
  borderline east={0.5pt}{0pt}{guidanceBoxAccent}
}
\renewenvironment{abstract}{%
  \begin{tcolorbox}[
    enhanced,
    breakable,
    colback=abstractBoxBg,
    boxrule=0pt,
    sharp corners,
    left=1.15em,
    right=1.15em,
    top=0.75em,
    bottom=0.8em,
    before skip=0pt,
    after skip=1.15em,
    borderline west={3pt}{0pt}{abstractBoxAccent}
  ]
  \noindent{\normalsize\bfseries\textcolor{abstractBoxAccent}{Abstract}}\\[0.35em]
  \small\setlength{\parindent}{0pt}\setlength{\baselineskip}{10.8pt}
}{%
  \end{tcolorbox}
}
\usepackage[most]{tcolorbox}
\usepackage{xcolor}

\definecolor{conclusionred}{RGB}{151,46,36}
\definecolor{conclusionbg}{RGB}{249,247,247}

\newtcolorbox{mainconclusion}{
    enhanced,
    breakable,
    colback=conclusionbg,
    frame hidden,
    boxrule=0pt,
    borderline west={2.2pt}{0pt}{conclusionred},
    borderline east={0.7pt}{0pt}{conclusionred},
    arc=0pt,
    left=10pt,
    right=10pt,
    top=7pt,
    bottom=7pt
}

\fancypagestyle{firstpage}{%
  \fancyhf{}%
  \fancyhead[L]{%
    \LARGE
    \begin{CJK*}{UTF8}{gbsn}\textcolor{projectLinkBlue}{Contribute New Tasks：}\end{CJK*}%
    \href{\taskSubmissionUrl}{\textcolor{projectLinkBlue}{\bfseries https://asibench.apexin.ai/submit}}%
  }%
}

\title{\texorpdfstring{
\resizebox{0.96\textwidth}{!}{%
\asiTitleA\asiTitleS\asiTitleI\textbf{-Bench:} At the Dawn of
\asiTitleA{}rtificial \asiTitleS{}uper\asiTitleI{}ntelligence%
}
}{ASI-Bench: At the Dawn of Artificial Superintelligence}}

\vspace{-13pt}
\author{
\begin{minipage}{0.95\textwidth}
\centering
\normalfont
\projectlink{\projectWebsiteUrl}{\faGlobe}{Website}
\textcolor{gray!55}{\quad\textbar\quad}
\projectlink{\projectGithubUrl}{\faGithub}{GitHub}
\textcolor{gray!55}{\quad\textbar\quad}
\projectlink{\projectHuggingFaceUrl}{\faDatabase}{HuggingFace}
\textcolor{gray!55}{\quad\textbar\quad}
\projectlink{\projectLeaderboardUrl}{\faTrophy}{Leaderboard}\\[1.45em]
{\normalsize\textbf{Core Authors}}\\[0.3em]
\small\setlength{\baselineskip}{11.8pt}
Junwei Zhou$^{5,\dagger}$, Zhen Sun$^{1,\dagger}$, Binyu Li$^{1}$,
Jiangyu Zhou$^{1}$, Yuexi Pan$^{1}$, Hengyu Wang$^{1}$,
Honghe Ren$^{1}$, Xiaohan Jia$^{1}$, \\ Xueyang Zhou$^{1}$, Xiaoyu Cao$^{1}$,
Yongchao Chen$^{1,*}$\\[0.12em]
\scriptsize $^{\dagger}$Equal contribution. \quad $^{*}$Corresponding author.\\[1.25em]
{\normalsize\textbf{Contributors}}\\[0.3em]
\small\setlength{\baselineskip}{11.8pt}
Yuanning Feng$^{1}$, Junhao Wu$^{1}$,
Cheng Zhang$^{13}$, Sijia Chen$^{10}$, Haoyu Xue$^{1}$,
Chengsong You$^{1}$,
Huan Wang$^{1}$, Koutian Wu$^{13}$,
Peigan Gao$^{9}$, Jiakun Wu$^{1}$,
Wenzhe Li$^{1}$, Ergan Shang$^{4}$, Qingyuan Zheng$^{1}$,
Jingjing Zhou$^{1}$, Ruixuan Jia$^{1}$, Yan Xu$^{2}$,
Hongrui Zhang$^{7}$, Xiao-Han Ma$^{9}$, Zhengxiang Cheng$^{1}$,
Yuexing Hao$^{2}$, Liting Mai$^{6}$, Xianglin Ji$^{2}$,
Wenjun Zhang$^{8}$, Zhuofan Chen$^{1}$, Yixiao Huang$^{1}$,
Chi Wang$^{12}$, Wenyue Hua$^{11}$, Yilun Hao$^{2}$,
Yuantao Zhai$^{1}$, Ziyan Zhao$^{1}$, Jingyan Xie$^{3}$\\[1.2em]
\scriptsize\setlength{\baselineskip}{9pt}
$^{1}$Tsinghua University \quad
$^{2}$Massachusetts Institute of Technology \quad
$^{3}$Harvard University \quad
$^{4}$Carnegie Mellon University \quad
$^{5}$University of Michigan \\
$^{6}$University of Illinois Urbana--Champaign \quad
$^{7}$Boston University \quad
$^{8}$University of Queensland \quad
$^{9}$University of Science and Technology of China \\
$^{10}$Flatiron Institute \quad
$^{11}$Microsoft Research \quad
$^{12}$AG2 AI \quad
$^{13}$Independent Researcher
\end{minipage}
}

\begin{document}
\setlength{\headheight}{22pt}
\maketitle
\thispagestyle{firstpage}
\enlargethispage{0.9in}
\vspace{-0.5in}

\begin{figure}[!ht]
  \centering
  \includegraphics[width=\linewidth]{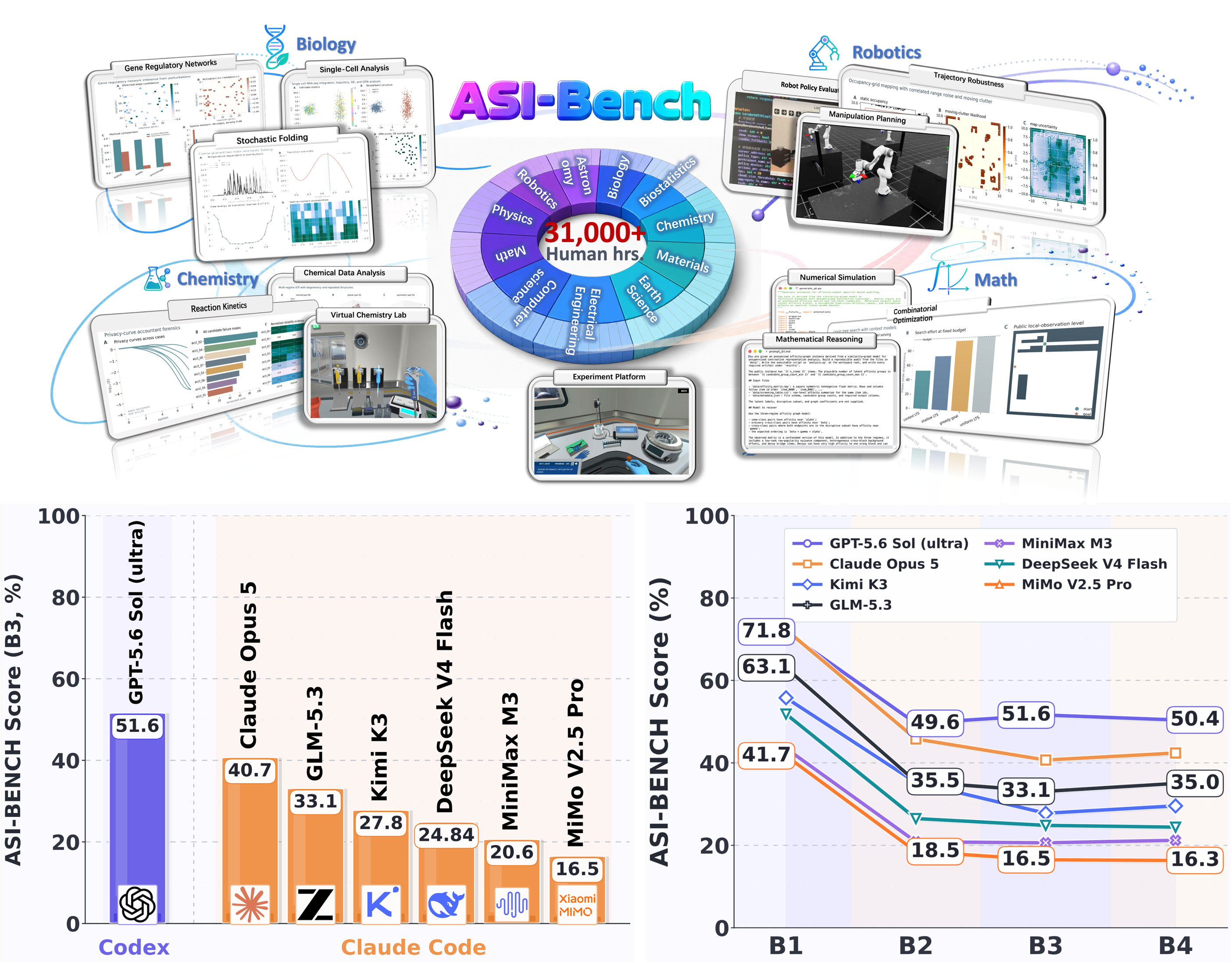}
  \caption{Overview of \asibench. Left: B3 performance across agents. Right: scores from B1 to B4, where B1 provides full methods, B2 only the method name, B3 only the research goal and data, and B4 further adds distractors.}
  \label{fig:benchmark-overview}
\end{figure}

\clearpage

\begin{abstract}

Artificial superintelligence (ASI) requires AI to move beyond mastering existing knowledge toward exploring the unknown, creating new knowledge, and turning new ideas into verifiable results. However, the capabilities of today’s AI systems are still largely built on learning, compressing, and applying existing human knowledge. Accordingly, existing benchmarks primarily test whether AI can produce correct answers based on learned knowledge, or whether it can complete tasks under extensive human guidance. We therefore introduce \asibench, the first benchmark to jointly evaluate AI systems' capabilities of innovative exploration and autonomous scientific execution across general research domains, and the first to progressively withdraw human methodological guidance within the same research project to test how far AI can proceed on its own. Built by over 40 experts with the cost of 31,000+ human hours, \asibench contains 60 project-level research tasks across 11 scientific domains and progressively reduces methodological guidance to test whether AI can independently select methods, conduct research, and produce verifiable results. All tasks undergo expert review, AI-assisted auditing, sandbox execution, and scorer validation. Across 18 state-of-the-art agent--model configurations, the average score drops from 50.91 with full methodological guidance to 29.10 with only the method specified and 26.62 when agents must determine the method themselves. This sharp decline shows that current systems remain heavily dependent on human guidance and are still far from autonomously conducting end-to-end, project-level scientific research. \asibench is open to the world. We invite researchers and builders everywhere to contribute new tasks, challenge the limits of today’s AI, and help accelerate humanity’s collective path toward artificial superintelligence at \url{https://asibench.apexin.ai/submit}.

\end{abstract}

\section{Introduction}
\label{sec:introduction}

A central challenge on the path toward artificial superintelligence (ASI) is whether AI can move beyond mastering existing human knowledge to explore unfamiliar problems, develop new solutions, and turn them into verifiable results. Today's AI systems derive much of their capability from learning, compressing, and applying the accumulated knowledge of humanity, and have made rapid progress in scientific reasoning, coding, data analysis, and agentic execution~\cite{Hua23b,Li24b,Sch25,Yua25,Pha25,Den25,Mer26}. Yet existing evaluations largely test these capabilities either through problems with known answers or through tasks whose methods and procedures are substantially specified by humans. They therefore provide limited evidence about whether AI can autonomously conduct scientific research when both the problem and the path to a solution are open-ended. In this work, we ask a more direct question: how far can current AI systems independently explore and execute project-level scientific research as human methodological guidance is progressively withdrawn?

\begin{wrapfigure}{r}{0.52\textwidth}
  \vspace{-1.0em}
  \centering
  \includegraphics[width=\linewidth]{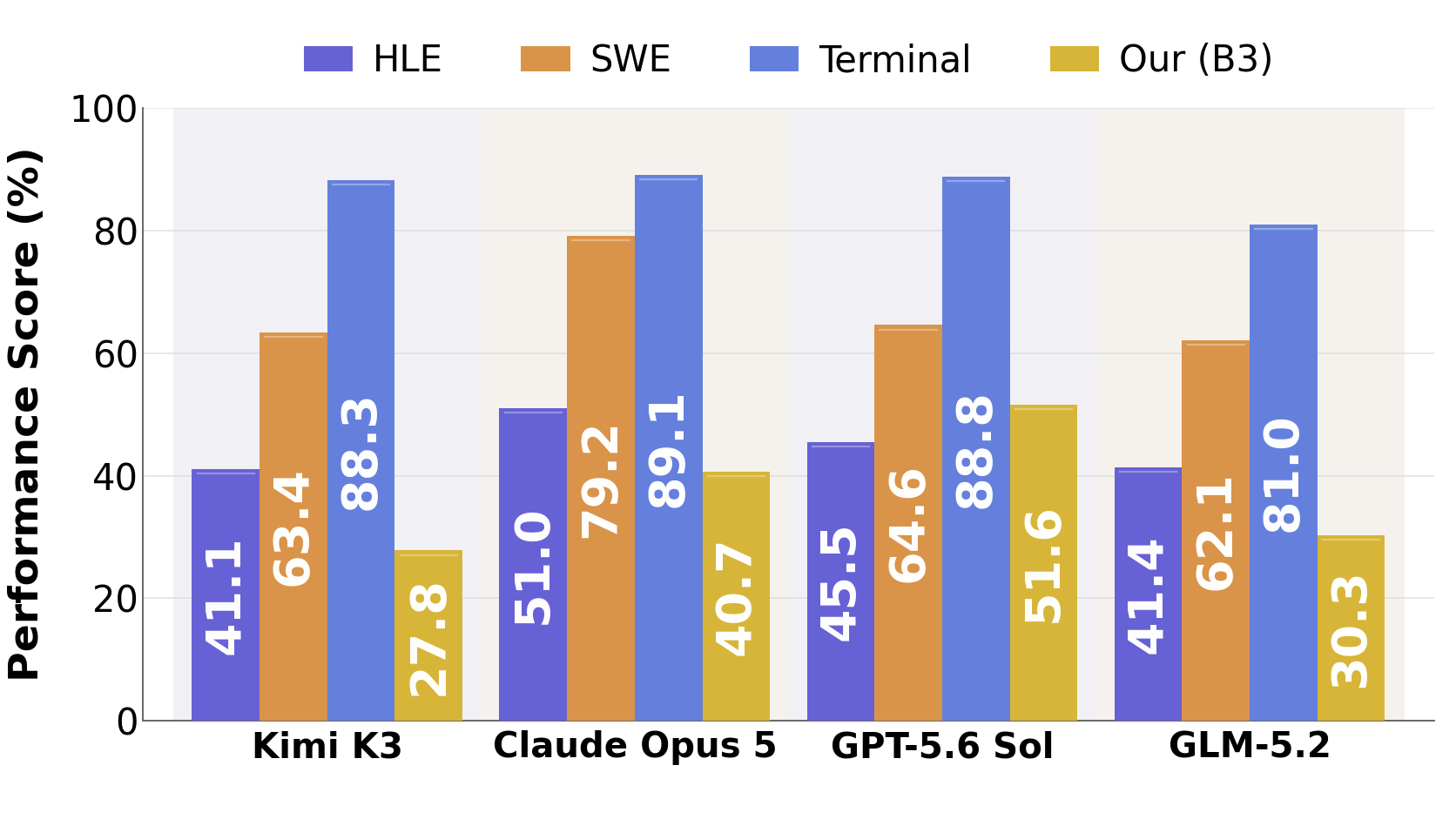}
  \caption{Performance comparison across HLE, SWE-bench, Terminal-Bench, and \asibench.}
  \label{fig:performance-compare}
  \vspace{-12pt}
\end{wrapfigure}

Built by over 40 experts with the cost of 31,000+ human hours, \asibench is designed to make this question directly measurable. It consists of 60 project-level research tasks across 11 scientific domains, with each task built around a scientific objective, task-specific data, an executable environment, and verifiable research artifacts. To measure autonomy, \asibench progressively reduces methodological guidance while keeping the research problem and evaluation criteria unchanged. In \textbf{B1}, the system receives complete methodological guidance; in \textbf{B2}, only the method is specified; in \textbf{B3}, the system must determine the method independently; and \textbf{B4} introduces task-irrelevant information under the B3 setting to evaluate robustness. This design distinguishes following a complete procedure, translating a specified method into an executable workflow, and independently conducting the research, enabling a unified assessment of general intelligence, innovation, and autonomous execution.

We employ a multi-stage construction and validation process to ensure the scientific validity and evaluation reliability of \asibench. Candidate research problems are first collected from traceable scientific sources, then screened and engineered into executable tasks by domain experts. Each task subsequently undergoes AI-assisted auditing and four rounds of human cross-review, covering scientific logic, task specifications, reference artifacts, scoring code, information leakage, and agent trajectories. Finally, every task is executed end-to-end in a sandbox environment to verify its reference solution and scoring behavior; tasks with scientific inconsistencies, unreliable evaluation, or unintended shortcuts are revised or excluded.

Across 18 state-of-the-art Agent$\times$Model configurations, average performance drops from 50.91 with full methodological guidance to 29.10 when only the method is specified and 26.62 when agents must determine the method themselves, revealing a substantial gap between scientific execution and autonomous discovery. 

\asibench provides a common reference point for measuring the transition from AI systems that primarily learn, compress, and apply existing human knowledge to systems capable of general intelligence, innovation, and autonomous execution. By revealing how far AI systems can progress as methodological guidance is withdrawn, \asibench establishes a benchmark for tracking the emergence of capabilities that may define the dawn of artificial superintelligence.

\section{Benchmark Design}
\label{sec:benchmark-design}


\newcolumntype{L}[1]{>{\raggedright\arraybackslash}p{#1}}
\newcolumntype{Z}{>{\centering\arraybackslash}X}

\begin{table*}[t]
\centering
\caption{Comparison of representative benchmarks for evaluating capabilities toward artificial superintelligence.}
\label{tab:benchmark_comparison}

\begin{threeparttable}

\small
\setlength{\tabcolsep}{4.2pt}
\renewcommand{\arraystretch}{1.14}

\begin{tabularx}{\textwidth}{
    @{}
    L{3.05cm}
    L{3.05cm}
    Z
    Z
    Z
    Z
    @{}
}

\specialrule{1.20pt}{0pt}{0pt}

\multirow{2}{*}{\textbf{Benchmark}}
&
\multirow{2}{*}{\textbf{Task Setting}}
&
\multicolumn{4}{c}{\textbf{Capabilities toward Autonomous Research}}
\\

\cmidrule[0.70pt](lr){3-6}

&
&
\makecell[c]{\bfseries\itshape Cross-domain\\ \bfseries\itshape Generality}
&
\makecell[c]{\bfseries\itshape Method\\ \bfseries\itshape Autonomy}
&
\makecell[c]{\bfseries\itshape End-to-end\\ \bfseries\itshape Research}
&
\makecell[c]{\bfseries\itshape Guidance\\ \bfseries\itshape Gradient}
\\

\specialrule{0.85pt}{0pt}{1pt}

Humanity's Last Exam~\cite{Pha25}
& Academic QA
& \checkmark
& --
& --
& --
\\[0.10em]

Terminal-Bench~\cite{Mer26}
& Terminal tasks
& --
& \(\triangle\)
& \(\triangle\)
& --
\\[0.10em]

ScienceAgentBench~\cite{Che24b}
& Scientific analysis
& \(\triangle\)
& \(\triangle\)
& \(\triangle\)
& --
\\[0.10em]

PaperBench~\cite{Sta25}
& Research replication
& --
& --
& \checkmark
& --
\\[0.10em]

MLE-Bench~\cite{Cha24}
& ML engineering
& --
& \(\triangle\)
& \checkmark
& --
\\[0.10em]

RE-Bench~\cite{Wij24}
& AI R\&D
& --
& \checkmark
& \checkmark
& --
\\[0.10em]

DiscoveryBench~\cite{Maj24}
& Scientific discovery
& \(\triangle\)
& \checkmark
& \(\triangle\)
& --
\\[0.10em]

SciCode~\cite{Tia24b}
& Scientific coding
& \checkmark
& \(\triangle\)
& --
& --
\\

\specialrule{0.85pt}{1.5pt}{0pt}

\rowcolor{black!5}
\textbf{ASI-Bench}
& \textbf{Project-level research}
& \textbf{\checkmark}
& \textbf{\checkmark}
& \textbf{\checkmark}
& \textbf{\checkmark}
\\

\specialrule{1.20pt}{0pt}{0pt}

\end{tabularx}

\vspace{3pt}

\begin{tablenotes}[flushleft]
\footnotesize
\item
\(\checkmark\) denotes explicit coverage;
\(\triangle\) denotes partial or domain-limited coverage;
-- denotes that the capability is not explicitly evaluated.
\end{tablenotes}

\end{threeparttable}
\end{table*}

\subsection{Why Do We Need \asibench?}

Existing benchmarks have pushed different dimensions of advanced AI capability substantially forward. Humanity's Last Exam (HLE)~\cite{Pha25} tests frontier knowledge across a broad range of disciplines, while SciCode~\cite{Tia24b} focuses on research-level scientific coding. Long-horizon agent benchmarks such as Terminal-Bench~\cite{Mer26} evaluate whether agents can use tools, resolve errors, and complete complex tasks. These benchmarks provide challenging tests of knowledge and execution, but typically operate with predefined objectives and evaluation criteria.

Other benchmarks focus on different components of scientific research. ScienceAgentBench~\cite{Che24b} evaluates data-driven scientific analysis, DiscoveryBench~\cite{Maj24} hypothesis-driven discovery, MLE-Bench~\cite{Cha24} machine-learning engineering, and RE-Bench~\cite{Wij24} open-ended AI R\&D; PaperBench~\cite{Sta25} focuses on research replication and computational reproducibility. As summarized in Table~\ref{tab:benchmark_comparison}, these benchmarks cover different combinations of cross-domain generality, methodological autonomy, and end-to-end execution, each capturing an important dimension of advanced intelligence.

What remains missing is a benchmark that evaluates these dimensions jointly. \asibench uses cross-domain project-level research to evaluate general intelligence, independent method selection to evaluate innovation, and end-to-end completion to evaluate autonomous execution. Its B1--B3 guidance gradient further measures whether these capabilities persist as human methodological guidance is progressively withdrawn. Together, these dimensions provide a more complete evaluation of intelligence than any individual capability alone.

\subsection{How Is ASI-Bench Designed?}

\asibench consists of 60 project-level research tasks across 11 scientific domains, designed to evaluate how far AI can conduct scientific research as human methodological guidance is progressively withdrawn. Each project is evaluated under matched guidance conditions while keeping the underlying task, data, required outputs, and scoring criteria fixed.


\paragraph{Autonomous Research with Reduced Human Guidance.}
Each task in \asibench is designed as a complex, project-level scientific investigation rather than an isolated question or a task with a predefined solution path. Starting from a research objective and task-specific data, agents must carry out a long-horizon, multi-stage research process that spans problem understanding, method selection, implementation, experimentation, failure diagnosis, iterative refinement, and result validation, ultimately producing verifiable scientific artifacts. Across the 60 tasks, completing these research processes involves more than 2,600 interaction turns and 2,400 execution steps, spanning over 35 hours of agent execution. Beyond their length, the tasks contain multiple interdependent and open-ended decision points: agents may need to choose among alternative methods, interpret intermediate results, recover from failed attempts, and revise subsequent actions accordingly. \asibench therefore evaluates sustained end-to-end scientific investigation rather than isolated question answering or the execution of a fixed procedure.

To measure scientific autonomy, \asibench progressively reduces human methodological guidance while keeping the research objective, data, and evaluation criteria fixed. A representative example is provided in Appendix~\ref{app:guidance-case-study}, where the same nonlinear two-dimensional dynamical system task is presented under B1--B4. In \textbf{B1}, the governing PDE, numerical formulation, and solver procedure are explicitly provided, where the agent mainly needs to implement and execute the prescribed approach. In \textbf{B2}, the full procedure is removed. The agent is only given methodological guidance about the class of PDE and suitable numerical approaches, and must turn this information into a working solution. In \textbf{B3}, even this methodological guidance is removed. The agent receives only the observed spatio-temporal data, the scientific objective, and the required outputs. It must determine the underlying model, choose an appropriate numerical method, implement it, and validate the resulting prediction. \textbf{B4} retains the B3 setting but adds plausible yet task-irrelevant information, testing whether the agent can maintain its research direction under distraction. This progression shifts scientific responsibility from humans to AI, from executing a prescribed procedure to independently deciding how the research should be conducted.

\paragraph{Broad Scientific Coverage.}
\asibench contains 60 project-level research tasks spanning 11 scientific domains: mathematics, physics, chemistry, biology, astronomy, materials science, earth science, medicine and biostatistics, computer science, robotics, and electrical engineering. These domains cover fundamental science, life science, computing, and engineering, requiring the same Agent$\times$Model system to generalize across different data types, scientific methods, and validation criteria. This breadth tests whether autonomous research transfers across disciplines rather than remaining limited to a single domain or familiar workflow.

\begin{figure*}[t]
\centering
\includegraphics[width=\linewidth,page=5]{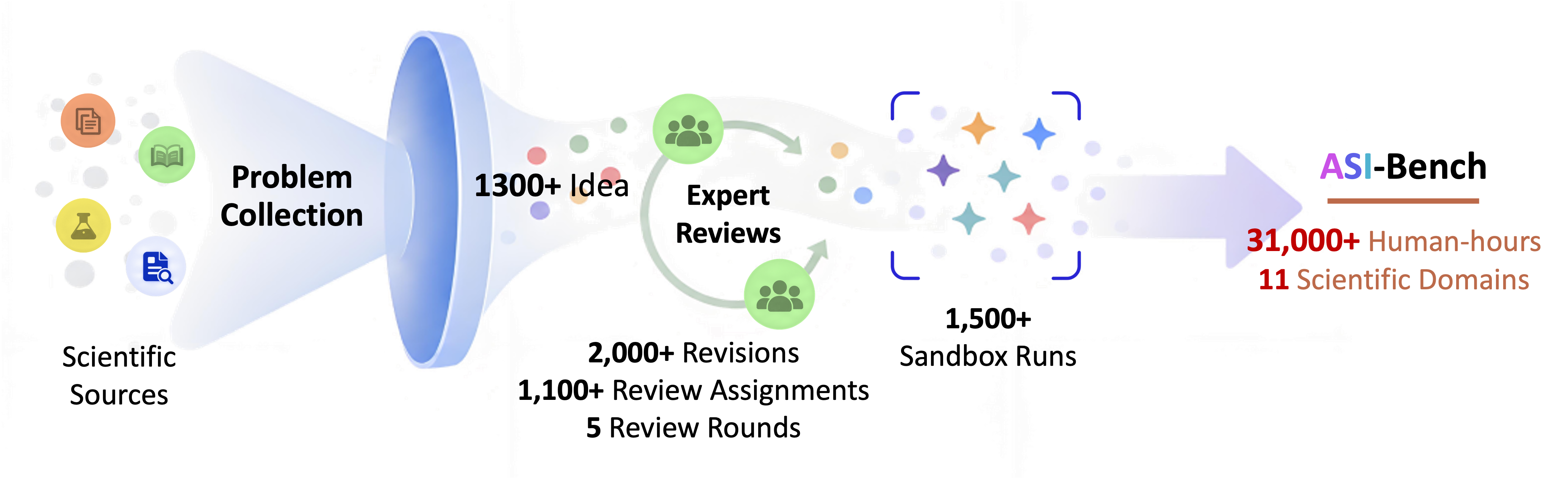}
\caption{
Representative project-level tasks in \asibench across physics, astronomy, electrical engineering, and computer science.
}
\label{fig:task-create_pipeline}
\vspace{-10pt}
\end{figure*}

\paragraph{Rigorous Construction and Validation.}
\asibench is built through a large-scale, iterative process rather than a single-pass collection. As shown in Figure~\ref{fig:task-create_pipeline}, the process begins with more than 1,300 candidate research ideas collected from scientific sources. These candidates are repeatedly reviewed, revised, or removed before entering the benchmark. The construction process includes five review rounds, more than 1,100 review assignments, and over 2,000 task revisions. Reviewers examine the scientific formulation, task specification, B1--B4 information design, reference results, and evaluation criteria. They also check for information leakage and unintended shortcuts that could lead to high scores without correctly solving the task. Across this construction and validation process, more than 31,000 human-hours were invested. Each retained task is further validated through end-to-end execution in isolated sandboxes. Across development, more than 1,500 sandbox runs are conducted to verify runtime stability, reference reproducibility, artifact generation, and scoring consistency. Tasks with unresolved scientific errors, unstable execution, evaluation misalignment, or unintended solution paths are revised or excluded. After this process, 60 project-level tasks spanning 11 scientific domains remain in the final benchmark.

\subsection{When Should \asibench Be Used?}

\textbf{To demonstrate stronger intelligence.}
Improvements of several percentage points on conventional benchmarks do not necessarily indicate progress in higher-order intelligence, especially when the problems, answers, or solution procedures are already well specified. Improvements on \asibench, particularly under low-guidance conditions where the system must determine how to approach the research problem itself, provide stronger evidence of progress in general intelligence, innovation, and autonomous execution. New foundation models, reasoning models, research agents, and general-purpose agents can therefore use \asibench to demonstrate advances beyond knowledge recall or procedure following.

\textbf{To test autonomous research.}
\asibench is particularly suitable for systems claiming capabilities in AI for Science, AI Scientist, automated research, or highly challenging tasks. By progressively removing methodological guidance, it tests whether a system can continue a research project when a complete human-designed procedure is no longer provided. The evaluation can further reveal whether failures arise from insufficient knowledge, method selection, tool execution, result interpretation, error correction, or stability.

\textbf{To compare models and agents.}
Different backbone models can be evaluated within the same agent framework, while the same model can be paired with different agents or harnesses to measure the contribution of system design. This allows researchers to distinguish improvements from the backbone model itself from those introduced by the agent design, and to identify under which guidance levels, scientific domains, and research stages the gains occur.

\textbf{To measure frontier progress.}
Current performance on \asibench remains low: across 18 state-of-the-art Agent$\times$Model configurations, the average B3 score is only 26.62. The benchmark is therefore far from saturated and retains substantial room to distinguish future systems. Significant and reproducible gains, particularly under minimal methodological guidance, can provide a clear signal of progress toward more autonomous and higher-order intelligence. \asibench can thus serve as a public evaluation and leaderboard for teams seeking to demonstrate such advances.

\textbf{To expand the benchmark together.}
\asibench is intended to grow with the scientific and AI communities. Researchers working at the frontier of different fields are often best positioned to identify research problems that are both scientifically meaningful and sufficiently challenging. We therefore invite scientists, engineers, model teams, agent developers, and benchmark researchers to contribute new problems, tasks, evaluation methods, and verified results, helping \asibench remain challenging and relevant as AI capabilities advance.

\section{Experiments}
\label{sec:experiments}

\subsection{Main Results}
\label{sec:main_results}

\begin{table*}[t]
\caption{
Main results on 60 project-level tasks without external tool access.
Scores are macro-averaged over tasks and, unless otherwise specified,
over three independent runs.
Gray rows report $\pm$ one sample standard deviation across independent runs.
The last three columns report B2$-$B1, B3$-$B1, and B4$-$B3, respectively,
with differences computed independently for each run before aggregation.
\textbf{Bold} values indicate the column-wise maximum result.
}
\label{tab:main-results-no-tools}

\centering
\small
\setlength{\tabcolsep}{3.8pt}
\renewcommand{\arraystretch}{0.90}

\begin{threeparttable}

\begin{tabularx}{\textwidth}{
@{}
p{1.75cm}
p{2.65cm}
Y
Y
Y
Y
Y
Y
Y
Y
@{}
}

\toprule[1.2pt]

\multirow{2}{*}{\textbf{Harness}}
&
\multirow{2}{*}{\textbf{Backbone Model}}
&
\multicolumn{4}{c}{
\cellcolor{black!5}
\textbf{Scientific Score}
}
&
\multirow{2}{*}{\textbf{Overall}}
&
\multicolumn{3}{c}{
\cellcolor{black!5}
\textbf{Diagnostic Metrics}
}
\\[-1pt]

\cmidrule[0.55pt](lr){3-6}
\cmidrule[0.55pt](lr){8-10}

&
&
\textbf{B1}
&
\textbf{B2}
&
\textbf{B3}
&
\textbf{B4}
&
&
\textbf{B2$-$B1}
&
\textbf{B3$-$B1}
&
\textbf{B4$-$B3}
\\

\midrule[0.8pt]


&
GPT-5.5 (xhigh)
&
57.57
&
35.28
&
29.29
&
30.46
&
38.15
&
$-22.29$
&
$-28.27$
&
$+1.16$
\\

\rowcolor{black!4}
\cellcolor{white}
&
\multicolumn{1}{l}{\scriptsize\color{black!55}\textit{$\pm$ std.}}
&
\multicolumn{1}{c}{\scriptsize\color{black!55}$\pm 5.74$}
&
\multicolumn{1}{c}{\scriptsize\color{black!55}$\pm 2.23$}
&
\multicolumn{1}{c}{\scriptsize\color{black!55}$\pm 1.57$}
&
\multicolumn{1}{c}{\scriptsize\color{black!55}$\pm 1.55$}
&
\multicolumn{1}{c}{\scriptsize\color{black!55}$\pm 1.25$}
&
\multicolumn{1}{c}{\scriptsize\color{black!55}$\pm 7.37$}
&
\multicolumn{1}{c}{\scriptsize\color{black!55}$\pm 5.26$}
&
\multicolumn{1}{c}{\scriptsize\color{black!55}$\pm 3.10$}
\\

&
GPT-5.6 Sol (xhigh)
&
62.75
&
42.96
&
40.86
&
40.74
&
46.83
&
$-19.79$
&
$-21.89$
&
$-0.12$
\\

\rowcolor{black!4}
\cellcolor{white}
&
\multicolumn{1}{l}{\scriptsize\color{black!55}\textit{$\pm$ std.}}
&
\multicolumn{1}{c}{\scriptsize\color{black!55}$\pm 3.05$}
&
\multicolumn{1}{c}{\scriptsize\color{black!55}$\pm 1.59$}
&
\multicolumn{1}{c}{\scriptsize\color{black!55}$\pm 2.35$}
&
\multicolumn{1}{c}{\scriptsize\color{black!55}$\pm 1.77$}
&
\multicolumn{1}{c}{\scriptsize\color{black!55}$\pm 1.10$}
&
\multicolumn{1}{c}{\scriptsize\color{black!55}$\pm 1.48$}
&
\multicolumn{1}{c}{\scriptsize\color{black!55}$\pm 5.27$}
&
\multicolumn{1}{c}{\scriptsize\color{black!55}$\pm 3.66$}
\\

&
GPT-5.6 Sol (ultra)
&
71.78
&
\textbf{49.57}
&
\textbf{51.60}
&
\textbf{50.41}
&
\textbf{55.84}
&
$-22.21$
&
$-20.18$
&
$-1.20$
\\

\rowcolor{black!4}
\cellcolor{white}
\multirow{-6}{*}{\textbf{Codex}}
&
\multicolumn{1}{l}{\scriptsize\color{black!55}\textit{$\pm$ std.}}
&
\multicolumn{1}{c}{\scriptsize\color{black!55}$\pm 0.46$}
&
\multicolumn{1}{c}{\scriptsize\color{black!55}$\pm 2.31$}
&
\multicolumn{1}{c}{\scriptsize\color{black!55}$\pm 3.65$}
&
\multicolumn{1}{c}{\scriptsize\color{black!55}$\pm 3.06$}
&
\multicolumn{1}{c}{\scriptsize\color{black!55}$\pm 1.58$}
&
\multicolumn{1}{c}{\scriptsize\color{black!55}$\pm 1.90$}
&
\multicolumn{1}{c}{\scriptsize\color{black!55}$\pm 3.21$}
&
\multicolumn{1}{c}{\scriptsize\color{black!55}$\pm 5.68$}
\\

\midrule[0.45pt]


&
Claude Opus 5$^{\dagger}$
&
\textbf{72.29}
&
45.80
&
40.70
&
42.39
&
50.29
&
$-26.49$
&
$-31.59$
&
$+1.69$
\\

\rowcolor{black!4}
\cellcolor{white}
&
\multicolumn{1}{c}{\scriptsize\color{black!55}\textit{single run}}
&
\multicolumn{1}{c}{\scriptsize\color{black!55}--}
&
\multicolumn{1}{c}{\scriptsize\color{black!55}--}
&
\multicolumn{1}{c}{\scriptsize\color{black!55}--}
&
\multicolumn{1}{c}{\scriptsize\color{black!55}--}
&
\multicolumn{1}{c}{\scriptsize\color{black!55}--}
&
\multicolumn{1}{c}{\scriptsize\color{black!55}--}
&
\multicolumn{1}{c}{\scriptsize\color{black!55}--}
&
\multicolumn{1}{c}{\scriptsize\color{black!55}--}
\\

&
Kimi K3
&
55.79
&
35.24
&
27.78
&
29.57
&
37.09
&
$-20.55$
&
$-28.02$
&
$+1.79$
\\

\rowcolor{black!4}
\cellcolor{white}
&
\multicolumn{1}{l}{\scriptsize\color{black!55}\textit{$\pm$ std.}}
&
\multicolumn{1}{c}{\scriptsize\color{black!55}$\pm 4.86$}
&
\multicolumn{1}{c}{\scriptsize\color{black!55}$\pm 3.36$}
&
\multicolumn{1}{c}{\scriptsize\color{black!55}$\pm 2.02$}
&
\multicolumn{1}{c}{\scriptsize\color{black!55}$\pm 0.46$}
&
\multicolumn{1}{c}{\scriptsize\color{black!55}$\pm 1.74$}
&
\multicolumn{1}{c}{\scriptsize\color{black!55}$\pm 3.88$}
&
\multicolumn{1}{c}{\scriptsize\color{black!55}$\pm 5.64$}
&
\multicolumn{1}{c}{\scriptsize\color{black!55}$\pm 2.33$}
\\

&
Claude Opus 4.8
&
52.48
&
36.25
&
31.73
&
29.59
&
37.51
&
$-16.24$
&
$-20.76$
&
$-2.14$
\\

\rowcolor{black!4}
\cellcolor{white}
&
\multicolumn{1}{l}{\scriptsize\color{black!55}\textit{$\pm$ std.}}
&
\multicolumn{1}{c}{\scriptsize\color{black!55}$\pm 6.27$}
&
\multicolumn{1}{c}{\scriptsize\color{black!55}$\pm 4.57$}
&
\multicolumn{1}{c}{\scriptsize\color{black!55}$\pm 3.44$}
&
\multicolumn{1}{c}{\scriptsize\color{black!55}$\pm 1.51$}
&
\multicolumn{1}{c}{\scriptsize\color{black!55}$\pm 0.79$}
&
\multicolumn{1}{c}{\scriptsize\color{black!55}$\pm 9.26$}
&
\multicolumn{1}{c}{\scriptsize\color{black!55}$\pm 8.71$}
&
\multicolumn{1}{c}{\scriptsize\color{black!55}$\pm 4.46$}
\\

&
GLM-5.3
&
63.05
&
35.45
&
33.09
&
35.01
&
41.65
&
$-27.60$
&
$-29.96$
&
$+1.92$
\\

\rowcolor{black!4}
\cellcolor{white}
&
\multicolumn{1}{l}{\scriptsize\color{black!55}\textit{$\pm$ std.}}
&
\multicolumn{1}{c}{\scriptsize\color{black!55}$\pm 1.40$}
&
\multicolumn{1}{c}{\scriptsize\color{black!55}$\pm 3.76$}
&
\multicolumn{1}{c}{\scriptsize\color{black!55}$\pm 1.36$}
&
\multicolumn{1}{c}{\scriptsize\color{black!55}$\pm 3.40$}
&
\multicolumn{1}{c}{\scriptsize\color{black!55}$\pm 1.35$}
&
\multicolumn{1}{c}{\scriptsize\color{black!55}$\pm 4.61$}
&
\multicolumn{1}{c}{\scriptsize\color{black!55}$\pm 1.30$}
&
\multicolumn{1}{c}{\scriptsize\color{black!55}$\pm 4.74$}
\\

&
GLM-5.2
&
54.01
&
30.81
&
30.29
&
27.27
&
35.59
&
$-23.20$
&
$-23.72$
&
$-3.02$
\\

\rowcolor{black!4}
\cellcolor{white}
&
\multicolumn{1}{l}{\scriptsize\color{black!55}\textit{$\pm$ std.}}
&
\multicolumn{1}{c}{\scriptsize\color{black!55}$\pm 1.65$}
&
\multicolumn{1}{c}{\scriptsize\color{black!55}$\pm 2.25$}
&
\multicolumn{1}{c}{\scriptsize\color{black!55}$\pm 1.96$}
&
\multicolumn{1}{c}{\scriptsize\color{black!55}$\pm 3.98$}
&
\multicolumn{1}{c}{\scriptsize\color{black!55}$\pm 0.95$}
&
\multicolumn{1}{c}{\scriptsize\color{black!55}$\pm 2.54$}
&
\multicolumn{1}{c}{\scriptsize\color{black!55}$\pm 3.56$}
&
\multicolumn{1}{c}{\scriptsize\color{black!55}$\pm 2.04$}
\\

&
DeepSeek V4 Flash
&
51.85
&
26.49
&
24.84
&
24.41
&
31.90
&
$-25.36$
&
$-27.01$
&
$-0.43$
\\

\rowcolor{black!4}
\cellcolor{white}
&
\multicolumn{1}{l}{\scriptsize\color{black!55}\textit{$\pm$ std.}}
&
\multicolumn{1}{c}{\scriptsize\color{black!55}$\pm 1.71$}
&
\multicolumn{1}{c}{\scriptsize\color{black!55}$\pm 1.24$}
&
\multicolumn{1}{c}{\scriptsize\color{black!55}$\pm 2.21$}
&
\multicolumn{1}{c}{\scriptsize\color{black!55}$\pm 0.95$}
&
\multicolumn{1}{c}{\scriptsize\color{black!55}$\pm 0.47$}
&
\multicolumn{1}{c}{\scriptsize\color{black!55}$\pm 0.82$}
&
\multicolumn{1}{c}{\scriptsize\color{black!55}$\pm 3.56$}
&
\multicolumn{1}{c}{\scriptsize\color{black!55}$\pm 1.28$}
\\

&
Kimi K2.7
&
44.43
&
23.84
&
19.75
&
21.34
&
27.34
&
$-20.59$
&
$-24.68$
&
$+1.58$
\\

\rowcolor{black!4}
\cellcolor{white}
&
\multicolumn{1}{l}{\scriptsize\color{black!55}\textit{$\pm$ std.}}
&
\multicolumn{1}{c}{\scriptsize\color{black!55}$\pm 1.90$}
&
\multicolumn{1}{c}{\scriptsize\color{black!55}$\pm 0.40$}
&
\multicolumn{1}{c}{\scriptsize\color{black!55}$\pm 0.74$}
&
\multicolumn{1}{c}{\scriptsize\color{black!55}$\pm 2.26$}
&
\multicolumn{1}{c}{\scriptsize\color{black!55}$\pm 0.61$}
&
\multicolumn{1}{c}{\scriptsize\color{black!55}$\pm 2.29$}
&
\multicolumn{1}{c}{\scriptsize\color{black!55}$\pm 2.07$}
&
\multicolumn{1}{c}{\scriptsize\color{black!55}$\pm 2.99$}
\\

&
MiniMax M3
&
43.53
&
20.86
&
20.61
&
21.21
&
26.55
&
$-22.68$
&
$-22.92$
&
$+0.60$
\\

\rowcolor{black!4}
\cellcolor{white}
&
\multicolumn{1}{l}{\scriptsize\color{black!55}\textit{$\pm$ std.}}
&
\multicolumn{1}{c}{\scriptsize\color{black!55}$\pm 3.60$}
&
\multicolumn{1}{c}{\scriptsize\color{black!55}$\pm 3.29$}
&
\multicolumn{1}{c}{\scriptsize\color{black!55}$\pm 1.02$}
&
\multicolumn{1}{c}{\scriptsize\color{black!55}$\pm 2.85$}
&
\multicolumn{1}{c}{\scriptsize\color{black!55}$\pm 1.55$}
&
\multicolumn{1}{c}{\scriptsize\color{black!55}$\pm 2.31$}
&
\multicolumn{1}{c}{\scriptsize\color{black!55}$\pm 3.20$}
&
\multicolumn{1}{c}{\scriptsize\color{black!55}$\pm 3.87$}
\\

&
DeepSeek V4 Pro
&
42.97
&
18.79
&
19.20
&
18.94
&
24.98
&
$-24.18$
&
$-23.77$
&
$-0.26$
\\

\rowcolor{black!4}
\cellcolor{white}
&
\multicolumn{1}{l}{\scriptsize\color{black!55}\textit{$\pm$ std.}}
&
\multicolumn{1}{c}{\scriptsize\color{black!55}$\pm 0.46$}
&
\multicolumn{1}{c}{\scriptsize\color{black!55}$\pm 2.88$}
&
\multicolumn{1}{c}{\scriptsize\color{black!55}$\pm 0.22$}
&
\multicolumn{1}{c}{\scriptsize\color{black!55}$\pm 1.23$}
&
\multicolumn{1}{c}{\scriptsize\color{black!55}$\pm 0.57$}
&
\multicolumn{1}{c}{\scriptsize\color{black!55}$\pm 2.61$}
&
\multicolumn{1}{c}{\scriptsize\color{black!55}$\pm 0.68$}
&
\multicolumn{1}{c}{\scriptsize\color{black!55}$\pm 1.28$}
\\

&
MiMo V2.5 Pro
&
41.70
&
18.49
&
16.49
&
16.33
&
23.25
&
$-23.22$
&
$-25.21$
&
$-0.16$
\\

\rowcolor{black!4}
\cellcolor{white}
\multirow{-20}{*}{\textbf{Claude Code}}
&
\multicolumn{1}{l}{\scriptsize\color{black!55}\textit{$\pm$ std.}}
&
\multicolumn{1}{c}{\scriptsize\color{black!55}$\pm 1.63$}
&
\multicolumn{1}{c}{\scriptsize\color{black!55}$\pm 2.27$}
&
\multicolumn{1}{c}{\scriptsize\color{black!55}$\pm 0.07$}
&
\multicolumn{1}{c}{\scriptsize\color{black!55}$\pm 0.39$}
&
\multicolumn{1}{c}{\scriptsize\color{black!55}$\pm 1.05$}
&
\multicolumn{1}{c}{\scriptsize\color{black!55}$\pm 1.20$}
&
\multicolumn{1}{c}{\scriptsize\color{black!55}$\pm 1.56$}
&
\multicolumn{1}{c}{\scriptsize\color{black!55}$\pm 0.31$}
\\

\midrule[0.45pt]


&
Kimi K3
&
56.16
&
34.05
&
28.15
&
26.53
&
36.22
&
$-22.11$
&
$-28.01$
&
$-1.62$
\\

\rowcolor{black!4}
\cellcolor{white}
&
\multicolumn{1}{l}{\scriptsize\color{black!55}\textit{$\pm$ std.}}
&
\multicolumn{1}{c}{\scriptsize\color{black!55}$\pm 4.86$}
&
\multicolumn{1}{c}{\scriptsize\color{black!55}$\pm 5.08$}
&
\multicolumn{1}{c}{\scriptsize\color{black!55}$\pm 0.84$}
&
\multicolumn{1}{c}{\scriptsize\color{black!55}$\pm 0.99$}
&
\multicolumn{1}{c}{\scriptsize\color{black!55}$\pm 2.22$}
&
\multicolumn{1}{c}{\scriptsize\color{black!55}$\pm 2.34$}
&
\multicolumn{1}{c}{\scriptsize\color{black!55}$\pm 4.56$}
&
\multicolumn{1}{c}{\scriptsize\color{black!55}$\pm 1.45$}
\\

&
Kimi K2.7
&
29.99
&
17.01
&
15.65
&
16.22
&
19.72
&
$\mathbf{-12.98}$
&
$\mathbf{-14.34}$
&
$+0.58$
\\

\rowcolor{black!4}
\cellcolor{white}
\multirow{-4}{*}{\textbf{Kimi Code}}
&
\multicolumn{1}{l}{\scriptsize\color{black!55}\textit{$\pm$ std.}}
&
\multicolumn{1}{c}{\scriptsize\color{black!55}$\pm 3.30$}
&
\multicolumn{1}{c}{\scriptsize\color{black!55}$\pm 0.10$}
&
\multicolumn{1}{c}{\scriptsize\color{black!55}$\pm 1.34$}
&
\multicolumn{1}{c}{\scriptsize\color{black!55}$\pm 0.85$}
&
\multicolumn{1}{c}{\scriptsize\color{black!55}$\pm 1.19$}
&
\multicolumn{1}{c}{\scriptsize\color{black!55}$\pm 3.35$}
&
\multicolumn{1}{c}{\scriptsize\color{black!55}$\pm 1.98$}
&
\multicolumn{1}{c}{\scriptsize\color{black!55}$\pm 1.56$}
\\

\midrule[0.45pt]


&
MiMo V2.5 Pro
&
27.73
&
11.33
&
11.50
&
14.11
&
16.17
&
$-16.40$
&
$-16.24$
&
$+2.61$
\\

\rowcolor{black!4}
\cellcolor{white}
\multirow{-2}{*}{\textbf{MiMo Code}}
&
\multicolumn{1}{l}{\scriptsize\color{black!55}\textit{$\pm$ std.}}
&
\multicolumn{1}{c}{\scriptsize\color{black!55}$\pm 4.33$}
&
\multicolumn{1}{c}{\scriptsize\color{black!55}$\pm 1.40$}
&
\multicolumn{1}{c}{\scriptsize\color{black!55}$\pm 1.36$}
&
\multicolumn{1}{c}{\scriptsize\color{black!55}$\pm 3.19$}
&
\multicolumn{1}{c}{\scriptsize\color{black!55}$\pm 1.39$}
&
\multicolumn{1}{c}{\scriptsize\color{black!55}$\pm 5.40$}
&
\multicolumn{1}{c}{\scriptsize\color{black!55}$\pm 5.37$}
&
\multicolumn{1}{c}{\scriptsize\color{black!55}$\pm 4.11$}
\\

\midrule[0.45pt]


&
DeepSeek V4 Flash
&
50.60
&
24.89
&
21.91
&
24.99
&
30.60
&
$-25.71$
&
$-28.69$
&
$\mathbf{+3.09}$
\\

\rowcolor{black!4}
\cellcolor{white}
&
\multicolumn{1}{l}{\scriptsize\color{black!55}\textit{$\pm$ std.}}
&
\multicolumn{1}{c}{\scriptsize\color{black!55}$\pm 2.24$}
&
\multicolumn{1}{c}{\scriptsize\color{black!55}$\pm 5.27$}
&
\multicolumn{1}{c}{\scriptsize\color{black!55}$\pm 1.75$}
&
\multicolumn{1}{c}{\scriptsize\color{black!55}$\pm 4.27$}
&
\multicolumn{1}{c}{\scriptsize\color{black!55}$\pm 3.05$}
&
\multicolumn{1}{c}{\scriptsize\color{black!55}$\pm 3.27$}
&
\multicolumn{1}{c}{\scriptsize\color{black!55}$\pm 1.53$}
&
\multicolumn{1}{c}{\scriptsize\color{black!55}$\pm 4.43$}
\\

&
DeepSeek V4 Pro
&
37.78
&
16.66
&
15.78
&
16.28
&
21.63
&
$-21.12$
&
$-22.00$
&
$+0.50$
\\

\rowcolor{black!4}
\cellcolor{white}
\multirow{-4}{*}{\textbf{OpenHands}}
&
\multicolumn{1}{l}{\scriptsize\color{black!55}\textit{$\pm$ std.}}
&
\multicolumn{1}{c}{\scriptsize\color{black!55}$\pm 2.28$}
&
\multicolumn{1}{c}{\scriptsize\color{black!55}$\pm 2.84$}
&
\multicolumn{1}{c}{\scriptsize\color{black!55}$\pm 1.91$}
&
\multicolumn{1}{c}{\scriptsize\color{black!55}$\pm 2.52$}
&
\multicolumn{1}{c}{\scriptsize\color{black!55}$\pm 0.75$}
&
\multicolumn{1}{c}{\scriptsize\color{black!55}$\pm 3.00$}
&
\multicolumn{1}{c}{\scriptsize\color{black!55}$\pm 3.42$}
&
\multicolumn{1}{c}{\scriptsize\color{black!55}$\pm 4.31$}
\\


\midrule[0.55pt]

\multicolumn{2}{l}{\textbf{All Models (Mean)}}
&
50.91
&
29.10
&
26.62
&
26.99
&
33.41
&
$-21.82$
&
$-24.29$
&
$+0.36$
\\

\bottomrule[1.2pt]

\end{tabularx}

\begin{tablenotes}[flushleft]
\footnotesize
\setlength{\itemsep}{0pt}
\setlength{\topsep}{3pt}

\item[]
Gray rows report $\pm$ one sample standard deviation across independent runs.
For the diagnostic metrics, differences are first computed within each run
and their standard deviations are then calculated across runs.

\item[$\dagger$]
This result is obtained from a single run; therefore, its standard deviation
cannot be estimated. All other results report the mean and standard deviation
over three independent runs.

\end{tablenotes}

\end{threeparttable}
\vspace{-12pt}
\end{table*}

We evaluate 18 representative Agent$\times$Model configurations on 60 tasks in \asibench. Table~\ref{tab:main-results-no-tools} reports their scores under B1--B4 together with the overall average.

\paragraph{Scientific Autonomy Remains Limited.}
Current systems remain far from reliable autonomous scientific discovery. Even the strongest configuration, Codex with GPT-5.6 Sol (ultra), achieves a B3 score of only 51.60 and is the sole evaluated system to exceed 50 when agents must independently select methods and construct research workflows. Stronger inference-time reasoning does improve this capability: increasing GPT-5.6 Sol from xhigh to ultra raises the B3 score from 40.86 to 51.60, a gain of 10.74 points. Yet this improvement also underscores how difficult scientific autonomy remains: substantial additional reasoning is required merely to reach moderate performance under reduced human guidance. The central challenge, therefore, is not simply whether AI can execute a research procedure once it is specified, but whether it can determine what procedure should be pursued in the first place. This distinction is critical for progress toward genuinely autonomous scientific research, where systems must move beyond following human-designed methodologies toward independently formulating, testing, and refining their own research strategies.

\paragraph{Dependence on Detailed Methodological Guidance.}
The results suggest that the main bottleneck is not method selection itself, but the ability to turn a scientific method into a complete research procedure. Average performance drops sharply from 50.91 in B1 to 29.10 in B2, a decrease of 21.82 points, when the detailed procedure is removed but the method remains available. By comparison, removing the method itself in B3 causes only a further 2.48-point drop, from 29.10 to 26.62. This large asymmetry indicates that current systems benefit far more from step-by-step methodological guidance than from being told which method to use. Performance remains nearly unchanged in B4 (26.99 versus 26.62 in B3), further showing that irrelevant contextual information has little effect relative to the loss of procedural guidance. Together, these results identify method operationalization, rather than method selection or distraction, as the primary bottleneck to autonomous scientific research.

\paragraph{Harnesses Shape Model's Capability.}
The harness can substantially change the capability expressed by the same underlying model. MiMo V2.5 Pro, for example, improves from 16.17 with MiMo Code to 23.25 with Claude Code, while Kimi K2.7 rises from 19.72 with Kimi Code to 27.34 with Claude Code. However, this effect is not uniform: Kimi K3 shows only a small difference between Kimi Code and Claude Code (36.22 vs.\ 37.09). These results show that scientific capability cannot be attributed to the backbone model alone; it emerges from the interaction between the model and the harness through which it reasons, executes, and organizes research. This has an important implication for both evaluation and system design: comparing models without accounting for their harnesses can obscure where capability actually comes from, while progress toward autonomous scientific research may depend as much on improving the surrounding research system as on scaling the model itself.

\subsection{Computational Cost}
\label{sec:cost-performance}

\begin{figure*}[t]
\centering
\includegraphics[width=\linewidth]{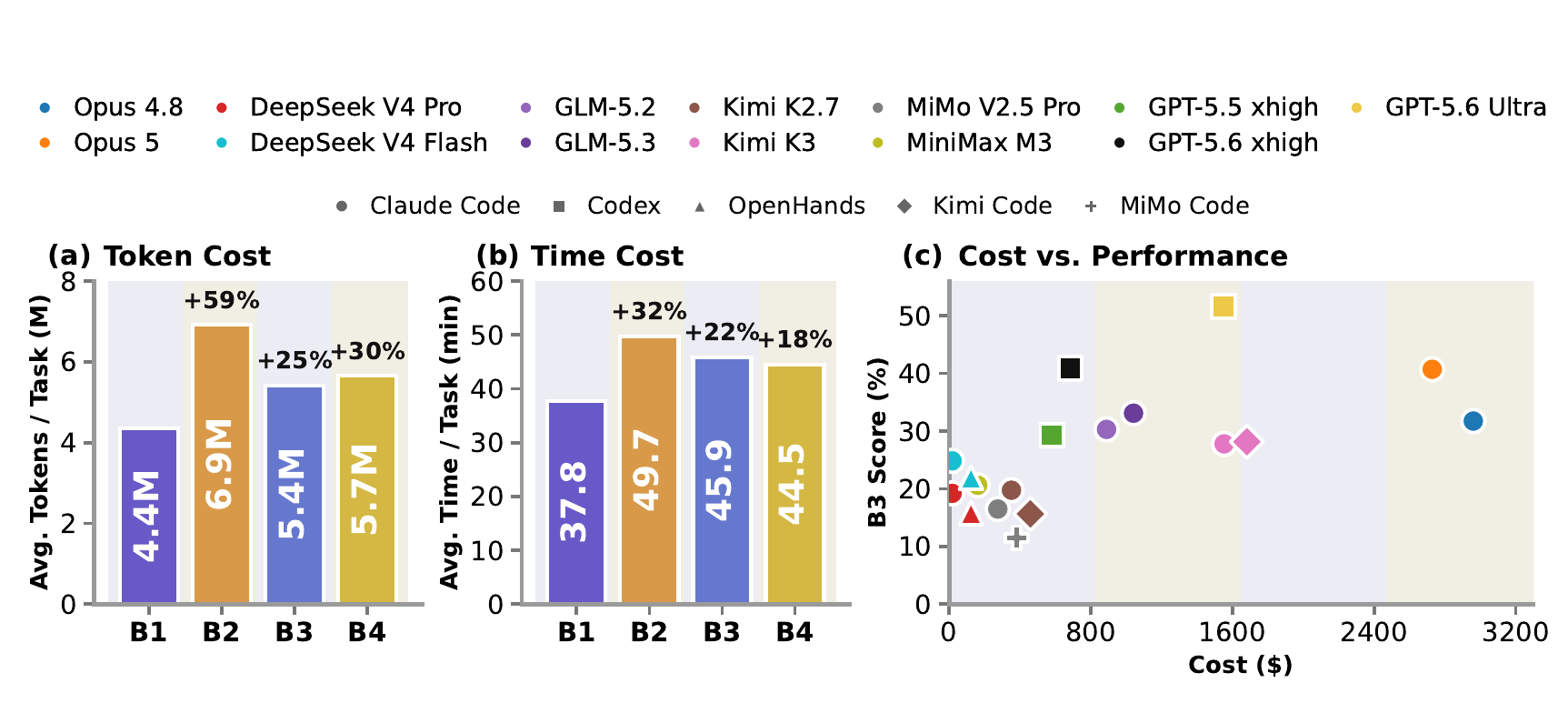} 
\caption{
Computational cost under different levels of methodological guidance and cost--performance trade-offs across Agent$\times$Model combinations.
(a--b) Average per-task token consumption and execution time under B1--B4.
(c) Relationship between per-run monetary cost and B3 scientific score across evaluated systems. Colors denote backbone models, while marker shapes denote agent harnesses.
}
\label{fig:cost-performance}
\vspace{-10pt}
\end{figure*}

Figure~\ref{fig:cost-performance} reports the average token consumption and execution time under different levels of methodological guidance, together with the cost profiles of individual Agent$\times$Model combinations.

\paragraph{Complete Guidance Reduces Compute Cost.}
Computational cost depends not only on how much methodological guidance is provided, but also on how complete that guidance is. B1, which specifies the method, implementation steps, and parameter choices, is the least expensive setting, requiring only 4.35M tokens and 37.8 minutes per task on average. Removing this procedural guidance increases the cost substantially: B3 and B4 consume 25\% and 30\% more tokens than B1 and require 22\% and 18\% more execution time, respectively, reflecting the additional exploration and iteration needed when agents must determine how to conduct the research themselves. More strikingly, B2 is the most expensive setting, requiring 6.91M tokens and 49.7 minutes per task---59\% more tokens and 32\% more time than B1. Providing only the method name therefore does not necessarily simplify research; instead, it can leave agents constrained by a prescribed direction while still requiring them to reconstruct the missing procedural details. This result reveals an important asymmetry in human guidance: complete guidance can sharply reduce search and implementation cost, whereas incomplete guidance may introduce additional overhead.

\paragraph{Higher Cost Does Not Guarantee Better Performance.}
Scientific performance varies substantially across configurations with similar or even very different computational budgets. Codex with GPT-5.6 xhigh achieves a B3 score of 40.86\% at approximately \$684 per run, closely matching Claude Opus 5 with Claude Code at 40.70\%, despite costing only about one quarter as much as its \$2,728 per-run cost. Spending more can still improve absolute performance: Codex with GPT-5.6 Ultra reaches the highest B3 score of 51.60\% at approximately \$1,550 per run. However, the large cost differences among configurations with comparable performance show that additional spending does not translate directly into stronger scientific capability. For practical autonomous research, system selection therefore involves a clear trade-off: GPT-5.6 xhigh offers the strongest cost--performance balance, whereas GPT-5.6 Ultra is preferable when maximizing scientific performance matters more than efficiency.

\section{Conclusion}
\label{sec:conclusion}

Beyond a single leaderboard, \asibench is intended to serve as shared research infrastructure for the scientific and AI communities. Its B1--B4 structure and domain-level analyses enable controlled comparisons across models and agents, while revealing where progress occurs and where human methodological guidance remains necessary. Yet no fixed benchmark—and no single research team—can fully represent the breadth of difficult, meaningful, and verifiable problems that future AI systems must confront. The continued value of \asibench therefore depends on collective participation from researchers working across disciplines and at the frontier of model development.

We invite scientists, engineers, model teams, agent researchers, and benchmark developers around the world to contribute research problems, executable tasks, evaluation methods, reviews, and verified results; to test new systems against the benchmark; and to challenge the benchmark itself as AI capabilities advance. By building \asibench as an open, rigorous, and continually evolving community standard, we can establish a shared foundation for identifying genuine advances in intelligence—and collectively accelerate progress toward artificial superintelligence.

\bibliographystyle{unsrt}
\bibliography{references}

 \clearpage
 \appendix

\appendix




\section{Related Work}
\label{sec:related-work}
\paragraph{AI Scientist Systems}

AI scientist systems combine language models with planning, retrieval, code execution, and scientific tools to automate increasingly complete research workflows \cite{Bra23,Boi23b,Lu24}. Coscientist demonstrated tool-augmented planning and experimentation in chemistry \cite{Boi23b}, while The AI Scientist integrated idea generation, implementation, experimentation, visualization, writing, and review in machine-learning research \cite{Lu24}. The AI Scientist-v2 further reduced its reliance on human-provided templates through experiment management and agentic tree search \cite{Yam25}. More recent systems, such as Co-Scientist and Kosmos, extend this direction toward collaborative hypothesis generation and long-horizon scientific investigation \cite{Eva26,Mit25}. Although these systems demonstrate broad workflow capabilities, their success does not necessarily indicate scientific autonomy, as objectives, methods, or experimental procedures may already be specified by humans \cite{Lu24,Lu26}. This motivates evaluations that distinguish autonomous workflow construction from reliable execution of provided methodology \cite{Che24b,Sta25}.

\paragraph{Benchmarks for Scientific Agents}

Scientific-agent evaluation has progressed from knowledge-based tests to executable research tasks. GPQA and Humanity's Last Exam assess advanced scientific knowledge without requiring data analysis or code execution \cite{Rei23,Pha25}. SciCode evaluates scientific programming \cite{Tia24b}, while DiscoveryBench, BLADE, DSBench, and HypoBench focus on data-driven discovery and hypothesis generation \cite{Maj24,Gu24,Jin24,Liu25}. These benchmarks target important capabilities but cover only selected stages of scientific research. More recent benchmarks evaluate longer workflows. ScienceAgentBench assesses data-driven scientific programming and execution \cite{Che24b}. MLE-Bench and RE-Bench evaluate open-ended machine-learning research engineering \cite{Cha24,Wij24}, while PaperBench, CORE-Bench, and ReplicationBench measure research replication through hierarchical artifact-based or execution-based evaluation \cite{Sta25,Sie24,Ye25}. FIRE-Bench and ResearchClawBench extend evaluation toward full-cycle rediscovery and end-to-end scientific research \cite{Wan26g,Xu26b}, and SciAgentArena introduces interactive cross-domain tasks with stepwise verification \cite{Liu26b}. DiscoveryWorld, AstaBench, and ScienceBoard further expand evaluation toward simulated, suite-based, and realistic multimodal scientific workflows \cite{Jan24,Bra25,Sun25}. These benchmarks improve realism, execution depth, and outcome verifiability. However, most provide each task under a fixed specification. Their scores therefore show whether an agent completes the assigned workflow, but not whether the workflow was independently constructed or derived from supplied methodology \cite{Maj24,Che24b,Sta25,Sie24}.

\paragraph{Scientific Autonomy and Guidance Dependence}

Several benchmarks vary the information available to a model. ScienceAgentBench compares performance with and without expert-provided knowledge, but its largely binary design does not distinguish method selection from procedural guidance \cite{Che24b}. ProjectionBench progressively reveals scientific information, yet focuses on hypothesis generation rather than executable project-level workflows \cite{Lew26}. ProjectEval varies specification detail for software projects \cite{Liu25c}, while contextual-robustness benchmarks examine sensitivity to irrelevant information. However, none evaluates how agents transition from following a prescribed method to independently constructing and validating a workflow within the same scientific project, or how this ability is affected by plausible distractors. ASI-Bench fills this gap through matched project instances that keep the research objective, data, required outputs, and evaluation criteria fixed while progressively withdrawing methodological guidance. Its four levels disentangle procedural execution, method identification, autonomous workflow construction, and distractor robustness, thereby directly measuring agents' dependence on human-provided methodology.

\section{Authors}
\label{app:authors}

\asibench is a collaborative benchmark built through broad participation in scientific task construction and independent expert review. The final benchmark contains 60 project-level research tasks spanning 11 scientific domains. In total, 21 researchers contributed tasks retained in the final benchmark. These tasks further underwent five rounds of human review, resulting in more than 1,100 task-review assignments and repeated revisions throughout benchmark construction. Task contributors are ordered primarily by the number of tasks retained in the final benchmark.

\subsection{Task Contributors}
\label{app:task-contributors}

The following researchers contributed scientific tasks retained in the final \asibench benchmark.

\noindent
Yuexi Pan$^{1}$,
Hengyu Wang$^{1}$,
Honghe Ren$^{1}$,
Peigan Gao$^{9}$,
Jiangyu Zhou$^{1}$,
Sijia Chen$^{10}$,
Junhao Wu$^{1}$,
Huan Wang$^{1}$,
Koutian Wu$^{13}$,
Cheng Zhang$^{13}$,
Yuanning Feng$^{1}$,
Qingyuan Zheng$^{1}$,
Wenzhe Li$^{1}$,
Jiakun Wu$^{1}$,
Ruixuan Jia$^{1}$,
Junwei Zhou$^{5}$,
Ergan Shang$^{4}$,
Jingjing Zhou$^{1}$,
Yan Xu$^{2}$,
Hongrui Zhang$^{7}$,
and Liting Mai$^{6}$.

\subsection{Human Reviewers}
\label{app:human-reviewers}

Across five rounds of review, more than 1,100 task-review assignments were completed. The following researchers participated in the human review process.

\noindent
Jiangyu Zhou$^{1}$,
Yuexi Pan$^{1}$,
Hengyu Wang$^{1}$,
Honghe Ren$^{1}$,
Xiaohan Jia$^{1}$,
Xueyang Zhou$^{1}$,
Cheng Zhang$^{13}$,
Yuanning Feng$^{1}$,
Sijia Chen$^{10}$,
Junhao Wu$^{1}$,
Huan Wang$^{1}$,
Koutian Wu$^{13}$,
Qingyuan Zheng$^{1}$,
Peigan Gao$^{9}$,
Wenzhe Li$^{1}$,
Jiakun Wu$^{1}$,
Jingjing Zhou$^{1}$,
Ruixuan Jia$^{1}$,
Yuexing Hao$^{2}$,
Yan Xu$^{2}$,
Hongrui Zhang$^{7}$,
Zhengxiang Cheng$^{1}$,
and Xianglin Ji$^{2}$.

\subsection{Affiliations}
\label{app:author-affiliations}

\vspace{0.4em}

\begin{multicols}{2}
\raggedcolumns
\footnotesize
\setlength{\columnsep}{2.5em}

\begin{enumerate}[
    label=\textsuperscript{\arabic*},
    leftmargin=1.7em,
    labelsep=0.35em,
    itemsep=0.25em,
    topsep=0pt,
    parsep=0pt
]

\item Tsinghua University, Beijing, China

\item Massachusetts Institute of Technology, Cambridge, MA, USA

\item Harvard University, Cambridge, MA, USA

\item Carnegie Mellon University, Pittsburgh, PA, USA

\item University of Michigan, Ann Arbor, MI, USA

\item University of Illinois Urbana--Champaign, Urbana, IL, USA

\item Boston University, Boston, MA, USA

\item The University of Queensland, Brisbane, QLD, Australia

\item University of Science and Technology of China, Hefei, China

\item Flatiron Institute, New York, NY, USA

\item Microsoft Research

\item AG2 AI

\item Independent Researcher

\end{enumerate}
\end{multicols}

\section{Contributing to \asibench}
\label{app:task-contribution}

\paragraph{Growing \asibench with the Research Community.}
\asibench is designed as an evolving, community-driven benchmark rather than a fixed collection of scientific problems. The first release contains 60 project-level tasks across 11 scientific domains, but these tasks represent only a small fraction of the scientific challenges on which increasingly capable AI systems should be evaluated. Many of the most meaningful problems are best identified by researchers working directly at the frontier of their respective fields. We therefore invite scientists, engineers, AI researchers, and benchmark developers to contribute new research tasks to future releases of \asibench. We particularly welcome problems that introduce new scientific domains, broaden the methodological diversity of existing domains, or test research capabilities that remain underrepresented in the current benchmark.

A contributed task should represent a genuine scientific problem rather than a conventional question-answering exercise. It should define a meaningful and reproducible scientific objective, require non-trivial reasoning or research execution, and produce artifacts that can be evaluated through explicit and reproducible criteria. Contributors retain the scientific substance of their research problems while expressing them through the standardized \asibench task format.

\begin{figure*}[t]
    \centering
    \includegraphics[width=\textwidth]{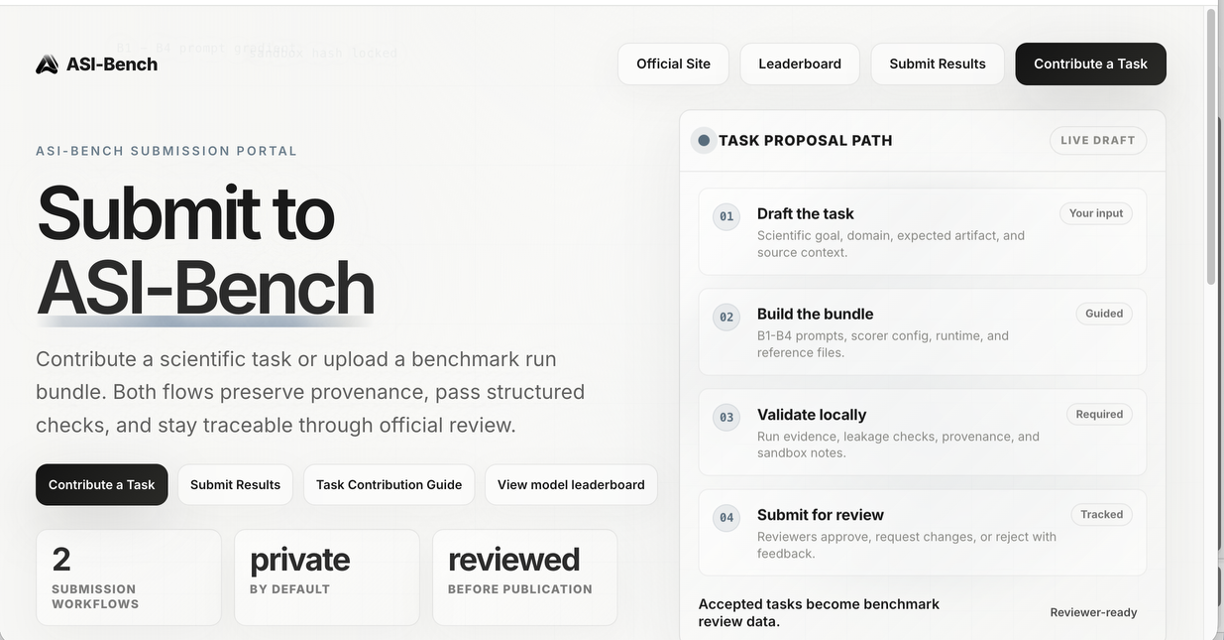}
    \caption{
    \textbf{\asibench task contribution portal.}
    The online submission workspace provides a 15-step Guided Flow for constructing and validating new tasks. Representative interfaces show the major stages of task authoring, including scientific problem definition, B1--B4 prompt construction, evaluation design with gates and scorers, runtime configuration, task-file preparation, and local testing before review.
    }
    \label{fig:task-submission-portal}
\end{figure*}

\paragraph{What Constitutes a Complete Task?}
A complete contribution contains not only a scientific question, but also the materials required to execute and evaluate it reproducibly. Each submission should include:
(i) a clearly stated scientific objective and an explanation of why the problem is non-trivial;
(ii) four prompt variants, B1--B4, defining the methodological-information gradient;
(iii) explicit input and output specifications for all agent-visible files and required artifacts;
(iv) a reproducible reference-generation procedure;
(v) an evaluation specification containing validity checks and weighted scoring criteria;
(vi) the scientific software dependencies required by the task; and
(vii) evidence from local evaluation across all four information conditions.

\paragraph{Constructing the B1--B4 Information Gradient.}
A central requirement of each contributed task is the controlled construction of four information conditions. \textbf{B1} provides the scientific background, method, equations, and procedural information required to execute the task. \textbf{B2} specifies the intended methodological approach and relevant constraints while leaving implementation decisions to the agent. \textbf{B3} specifies only the scientific objective, available inputs, constraints, and required outputs, requiring the agent to determine an appropriate solution strategy independently. \textbf{B4} extends B3 with factually correct but non-essential information, allowing robustness to irrelevant context to be evaluated. In particular, B3 and B4 should not reveal the intended algorithm, solver, method, or other information that would compromise the methodological gradient.

\paragraph{How to Contribute a Task.}
To make task contribution accessible across scientific disciplines, \asibench provides a public authoring scaffold and an online Guided Flow. As shown in Figure~\ref{fig:task-submission-portal}, contributors can start from the public task template and complete the submission through the online portal at \url{https://asibench.apexin.ai/submit} or via the CLI launcher. Before submission, each task must form a self-contained and reproducible package. It should include a well-defined scientific objective, four B1--B4 prompts, task data, a reference-generation script (\texttt{generate\_gt.py}), an evaluation configuration, runtime dependencies, and local-testing evidence. The contribution process consists of three main stages:

\begin{enumerate}[leftmargin=1.6em,itemsep=0.45em,topsep=0.35em]

    \item \textbf{Formulate the Scientific Question.}
Contributors first define the scientific question to be investigated and explain why it is non-trivial. The task should specify \emph{what} must be computed or discovered, rather than prescribing \emph{how} to solve it. It should also declare the scientific dependencies required by the task. The target result must be deterministic and reproducible so that it can support reliable evaluation.

    \item \textbf{Build the Four Prompt Levels.}
    Contributors construct B1--B4 for the same scientific objective. B1 provides the complete method and detailed methodological guidance. B2 retains method-level guidance but leaves implementation decisions to the agent. B3 specifies only the objective, inputs, constraints, and required artifacts, without revealing the intended method. B4 keeps the B3 task unchanged while adding factually correct but non-essential information. B3 and B4 must remain complete task specifications, including all required inputs and outputs. Contributors also provide \texttt{generate\_gt.py} to generate the reference result and define evaluation gates and weighted scorers for the required scientific outputs.

    \item \textbf{Test, Submit, and Revise.}
    Before submission, contributors evaluate B1--B4 using the model and evaluation settings specified by the submission portal at \url{https://asibench.apexin.ai/submit/}, and report the resulting scores. Local testing also records execution time, environment information, sandbox provenance, and scorer-replay evidence. These results are used to check task difficulty, information leakage, reproducibility, and evaluation stability. Once the required files and evidence are complete, the task can be submitted for review. Reviewers examine the scientific formulation, B1--B4 information gradient, reference generation, scoring logic, and runtime behavior. Contributors then revise the task in response to feedback until it satisfies the benchmark requirements.

\end{enumerate}

\paragraph{Contribution Recognition.}
Community contributors are treated as participants in the construction of \asibench rather than merely as external data submitters. For every contributed task that successfully passes scientific and technical review and is incorporated into an official \asibench release, the corresponding task contributor(s) will be included in the contributor list of that benchmark release.

\paragraph{An Open Benchmark for the Path Ahead.}
The current 60 tasks should be viewed as a starting point rather than a closed test set. The scientific problems that will distinguish increasingly capable AI systems cannot be defined by a single research group alone. We therefore invite researchers from different disciplines to contribute the research problems they believe future AI systems should be able to solve, and invite benchmark and model developers to help refine evaluations, identify failure cases, and validate new tasks. By continuously incorporating new scientific challenges from the research community, we aim for \asibench to evolve together with AI capability, challenge the limits of today's AI, and ultimately help measure and accelerate progress beyond the frontier of human scientific capability toward artificial superintelligence.

\section{Case Study}
\label{app:guidance-case-study}

\begin{taskbox}
\textbf{Task: 2D Anisotropic Stiff Dynamics}

\medskip

The agent is given observations of a two-dimensional nonlinear dynamical
system on a periodic spatial domain.

\medskip

\textbf{Input data}
\begin{itemize}
    \item \texttt{system\_info.json}: system parameters and timing metadata;
    \item \texttt{field\_evolution.npy}: observed spatio-temporal field snapshots;
    \item \texttt{initial\_condition.npy}: initial field for the prediction task.
\end{itemize}

\textbf{Scientific objective}

Given the observations, the agent must:
\begin{enumerate}
    \item characterize the spatial and temporal dynamics;
    \item construct a model that reproduces the observed dynamics;
    \item predict the field at a future target time;
    \item extract physically meaningful diagnostics.
\end{enumerate}

\textbf{Required artifacts}

The agent must produce a predicted field, spatial spectrum, physical
diagnostics, scientific visualizations, data-analysis results, and the
complete executable simulation code.

\medskip

The scientific objective, input data, required artifacts, and evaluation
criteria are kept fixed across B1--B4. Only the information provided in the
prompt changes.
\end{taskbox}


\begin{bpromptbox}
{B1 Prompt --- Full Equation and Solver Guide}
{b1blue}
{b1frame}

\small

\textbf{Task Overview}

You are given observed spatio-temporal evolution data from a 2D nonlinear
pattern-forming system governed by a conserved anisotropic
Kuramoto--Sivashinsky-type equation on a doubly periodic domain:

\begin{verbatim}
u_t = -alpha * u
      + u_xx + mu * u_yy
      - nu * u_xxxx
      - 2 * gamma * u_xxyy
      - delta * u_yyyy
      + nabla^2 [
          (lambda_xx / 2) * u_x^2
          + lambda_xy * u_x * u_y
          + (lambda_yy / 2) * u_y^2
        ]
\end{verbatim}

The domain is \texttt{[0,Lx] x [0,Ly]}. The prompt provides the values of
\texttt{Lx}, \texttt{Ly}, \texttt{alpha}, \texttt{nu}, \texttt{mu},
\texttt{gamma}, \texttt{delta}, \texttt{lambda\_xx},
\texttt{lambda\_xy}, and \texttt{lambda\_yy}.

\medskip

\textbf{Spatial Discretization}

Use a 2D Fourier pseudospectral method. Define

\begin{verbatim}
kx = 2*pi*fftfreq(Nx, d=dx)
ky = 2*pi*fftfreq(Ny, d=dy)
\end{verbatim}

and the Fourier-space linear operator

\begin{verbatim}
L_hat =
    -alpha
    + kx^2
    + mu * ky^2
    - nu * kx^4
    - 2 * gamma * kx^2 * ky^2
    - delta * ky^4
\end{verbatim}

Compute the nonlinear term pseudospectrally in conserved form:

\begin{verbatim}
f(u_x, u_y) =
    (lambda_xx / 2) * u_x^2
    + lambda_xy * u_x * u_y
    + (lambda_yy / 2) * u_y^2

N_hat = -(kx^2 + ky^2) * fft2(f)
\end{verbatim}

Apply 2/3-rule dealiasing to the nonlinear term.

\medskip

\textbf{Time Integration}

The system is stiff because of the fourth-order terms.
Use ETDRK4.

Precompute

\begin{verbatim}
E  = exp(L_hat * dt)
E2 = exp(L_hat * dt / 2)
\end{verbatim}

and construct the ETDRK4 coefficients using the
Kassam--Trefethen contour-integral formulas.

For the current Fourier state:

\begin{verbatim}
a = E2 * u_hat + Q * N_hat(u_hat)
b = E2 * u_hat + Q * N_hat(a)
c = E2 * a
    + Q * (2*N_hat(b) - N_hat(u_hat))

u_hat_next =
      E * u_hat
    + f1 * N_hat(u_hat)
    + 2*f2 * (N_hat(a) + N_hat(b))
    + f3 * N_hat(c)
\end{verbatim}

Your job is to analyze the observed data, reconstruct the numerical solver,
predict the field at the target time, and produce the required scientific
artifacts.

\end{bpromptbox}


\begin{bpromptbox}
{B2 Prompt --- Method Background}
{b2blue}
{b2frame}

\small

\textbf{Task}

You are given observed data from a 2D nonlinear evolution system that
produces directionally uneven, spatially complex patterns. The system is
stiff and exhibits irregular pattern dynamics rather than simple steady
diffusion.

Your goal is to analyze the data, build a numerical model that reproduces
the dynamics from the supplied initial condition, predict the future field,
and extract several physical diagnostics.

\medskip

\textbf{Physical and Numerical Background}

This is a fourth-order nonlinear PDE problem in two spatial dimensions.
The main ingredients are:

\begin{itemize}
    \item weak linear damping;
    \item destabilizing long-wave growth through second-order terms;
    \item stabilizing high-order dissipation through fourth-order terms,
    including cross-derivative coupling;
    \item nonlinear energy transfer with a conservation-type structure;
    \item different behavior in the two spatial directions.
\end{itemize}

Because of the fourth-order terms, the linear stiffness can be severe.
For this class of systems, methods based on spectral discretization with
stiffness-aware time stepping are needed.

Reasonable time-stepping approaches include:

\begin{itemize}
    \item ETD methods such as ETDRK4;
    \item IMEX schemes;
    \item semi-implicit spectral integrators.
\end{itemize}

ETD-style spectral solvers are often competitive, but they are not the only
reasonable option.

The system parameters are supplied in
\texttt{data/system\_info.json} with neutralized coefficient names.

\medskip

Inspect the observed data before locking in a solver. The two spatial
directions should be analyzed separately where appropriate. The task is not
only to forecast the field, but also to recover interpretable physical
structure from the data.

\end{bpromptbox}


\begin{bpromptbox}
{B3 Prompt --- Research Objective and Data Only}
{b2blue}
{b2frame}

\small

\textbf{Task}

You are given observed spatio-temporal data from a 2D nonlinear system on
a periodic domain. The field develops structured patterns and evolves in a
complex, nontrivial way over time.

\medskip

\textbf{Input Data}

\begin{itemize}
    \item \texttt{data/system\_info.json}: system parameters and timing metadata;
    \item \texttt{data/field\_evolution.npy}: observed 2D field snapshots;
    \item \texttt{data/initial\_condition.npy}: initial field used for prediction.
\end{itemize}

\textbf{Goal}

\begin{enumerate}
    \item Explore the observed data to characterize its spatial and temporal
    structure.
    \item Identify or construct a mathematical model that can reproduce the
    observed dynamics from the supplied initial condition.
    \item Predict the field at the target time specified in
    \texttt{system\_info.json}.
    \item Extract physically meaningful quantities from both the observed
    data and the predicted field.
\end{enumerate}

\textbf{Constraints}

\begin{itemize}
    \item Outputs must be finite, with no NaN or Inf values.
    \item Use the provided data files only.
    \item The governing equation, numerical representation, and effective
    evolution strategy are not given explicitly. Identifying an appropriate
    approach is part of the task.
\end{itemize}

\end{bpromptbox}


\begin{bpromptbox}
{B4 Prompt --- Research Objective with Distracting Information}
{b4blue}
{b4frame}

\small

\textbf{Task}

You are given observed spatio-temporal data from a 2D nonlinear system on
a periodic domain. The field develops structured patterns and evolves in a
complex, nontrivial way over time.

\medskip

\textbf{Goal}

\begin{enumerate}
    \item Explore the observed data to characterize its spatial and temporal
    structure.
    \item Identify or construct a mathematical model that can reproduce the
    observed dynamics.
    \item Predict the field at the specified target time.
    \item Extract physically meaningful quantities from the observed and
    predicted fields.
\end{enumerate}

\textbf{Additional Context}

Several unrelated 2D PDE families can produce structured fields with
evolving spectra. Reaction--diffusion systems such as FitzHugh--Nagumo or
Gray--Scott models generate Turing patterns through activator--inhibitor
interactions. Phase-field models such as Cahn--Hilliard describe spinodal
decomposition. Swift--Hohenberg-type equations model convective pattern
formation. Thin-film equations describe coating flows and droplet spreading.
Complex Ginzburg--Landau variants model oscillatory instabilities.

There are also many possible computational routes. Finite differences with
explicit stepping are straightforward. IMEX schemes treat different parts of
the operator differently. Exponential integrators handle the linear part
through matrix exponentials. Operator splitting divides the problem into
substeps. Pseudospectral solvers use FFTs. Neural operators attempt to learn
the dynamics directly from data.

For spatial characterization, possible approaches include Fourier analysis,
spatial autocorrelation, variograms, structure functions, wavelet
decomposition, proper orthogonal decomposition, and template matching.
Temporal structure can be analyzed through autocorrelation, recurrence
analysis, or temporal spectra.

\medskip

Some preliminary notes are also attached:

\begin{quote}
``maybe it is phase-field or Cahn--Hilliard?''

``could be some activator--inhibitor reaction--diffusion thing?''

``try a simpler neural surrogate first?''

``maybe just compare the last observed frame to the target frame?''

``check if the patterns are just equilibrium Turing patches''
\end{quote}

There is also unrelated background chatter about deadlines, news,
semiconductor markets, sports, entertainment, and upcoming meetings.

Your job remains the same: focus on the supplied scientific data, infer a
suitable model, produce the required outputs, and ignore irrelevant
information.

\end{bpromptbox}

\end{document}